\def\paperID{}
\def\confYear{2026}

\def\paperTitle{Video Generation with Stable Transparency via\\Shiftable RGB-A Distribution Learner}

\newif\ifreview 
\newif\ifarxiv \newcommand{\arxiv}{\arxivtrue}
\newif\ifcamera 
\newif\ifrebuttal 

\arxiv
\pdfoutput=1
\documentclass[10pt,twocolumn,letterpaper]{article}
\ifreview \usepackage[review]{cvpr} \fi
\ifarxiv \usepackage[pagenumbers]{cvpr} \fi
\ifrebuttal \usepackage[rebuttal]{cvpr} \fi
\ifcamera \usepackage{cvpr} \fi

\usepackage{graphicx}	
\usepackage{amsmath}	
\usepackage{amssymb}	
\usepackage{booktabs}
\usepackage{times}
\usepackage{microtype}
\usepackage{epsfig}
\usepackage{caption}
\usepackage{float}
\usepackage{placeins}
\usepackage{color, colortbl}
\usepackage{stfloats}
\usepackage{enumitem}
\usepackage{tabularx}
\usepackage{xstring}
\usepackage{multirow}
\usepackage{xspace}
\usepackage{url}
\usepackage{subcaption}
\usepackage{xcolor}
\usepackage[hang,flushmargin]{footmisc}
\usepackage[most]{tcolorbox}
\usepackage{pifont}
\usepackage{makecell}
\usepackage{threeparttable}
\usepackage[ruled,vlined]{algorithm2e}
\usepackage{tcolorbox}
\tcbuselibrary{skins,breakable}
\definecolor{myyellow}{HTML}{FDFCE3}
\definecolor{myframegray}{gray}{0.6}

\newtcolorbox{mytextbox}[1][]{
  arc=2mm,              
  colback=myyellow,       
  colframe=white,   
  boxrule=0.2pt,          
  boxsep=5pt,             
  left=5pt,              
  right=5pt,             
  top=3pt,                
  bottom=3pt,             
  breakable,              
  #1                      
}
\ifcamera \usepackage[accsupp]{axessibility} \fi

% ...

\ifarxiv  \fi

\newcommand{\R}[1]{{%
    \textbf{%
        \ifstrequal{#1}{1}{\textcolor{red}{R#1}}{%
        \ifstrequal{#1}{2}{\textcolor{blue}{R#1}}{%
        \ifstrequal{#1}{3}{\textcolor{magenta}{R#1}}{%
        \ifstrequal{#1}{4}{\textcolor{teal}{R#1}}{%
                           \textcolor{cyan}{R#1}%
        }}}}%
    }%
}}

\usepackage{xr-hyper}

\makeatletter
\newcommand*{\addFileDependency}[1]{
  \typeout{(#1)}
  \@addtofilelist{#1}
  \IfFileExists{#1}{}{\typeout{No file #1.}}
}

\makeatother

\definecolor{cvprblue}{rgb}{0.21,0.49,0.74}
\usepackage[pagebackref,breaklinks,colorlinks,allcolors=cvprblue]{hyperref}
\usepackage[capitalize]{cleveref}
\crefname{section}{Sec.}{Secs.}
\crefname{table}{Table}{Tables}
\crefname{figure}{Fig.}{Figs.}

\ifarxiv \crefname{appendix}{App.}{Apps.}
\else \crefname{appendix}{Suppl.}{Suppls.} \fi

\begin{document}
\title{\paperTitle}
\author{
    Haotian Dong$^{1}$ \quad Wenjing Wang$^{2}$ \quad Chen Li$^{2}$ \quad Jing Lyu$^{2}$ \quad Di Lin$^{1}$ \\[5pt]
    $^{1}$Tianjin University, China \qquad $^{2}$Independent Researcher\\
    {\tt\small htdong@tju.edu.cn, \{augustawang, chaselli, eckolv\}@tencent.com,} 
    {\tt\small ande.lin1988@gmail.com}
}

\twocolumn[{%
\renewcommand\twocolumn[1][]{#1}%
\maketitle
\begin{center}
    \centering
    \captionsetup{type=figure}
    \vspace{-5mm}
    \includegraphics[width=0.99\linewidth]{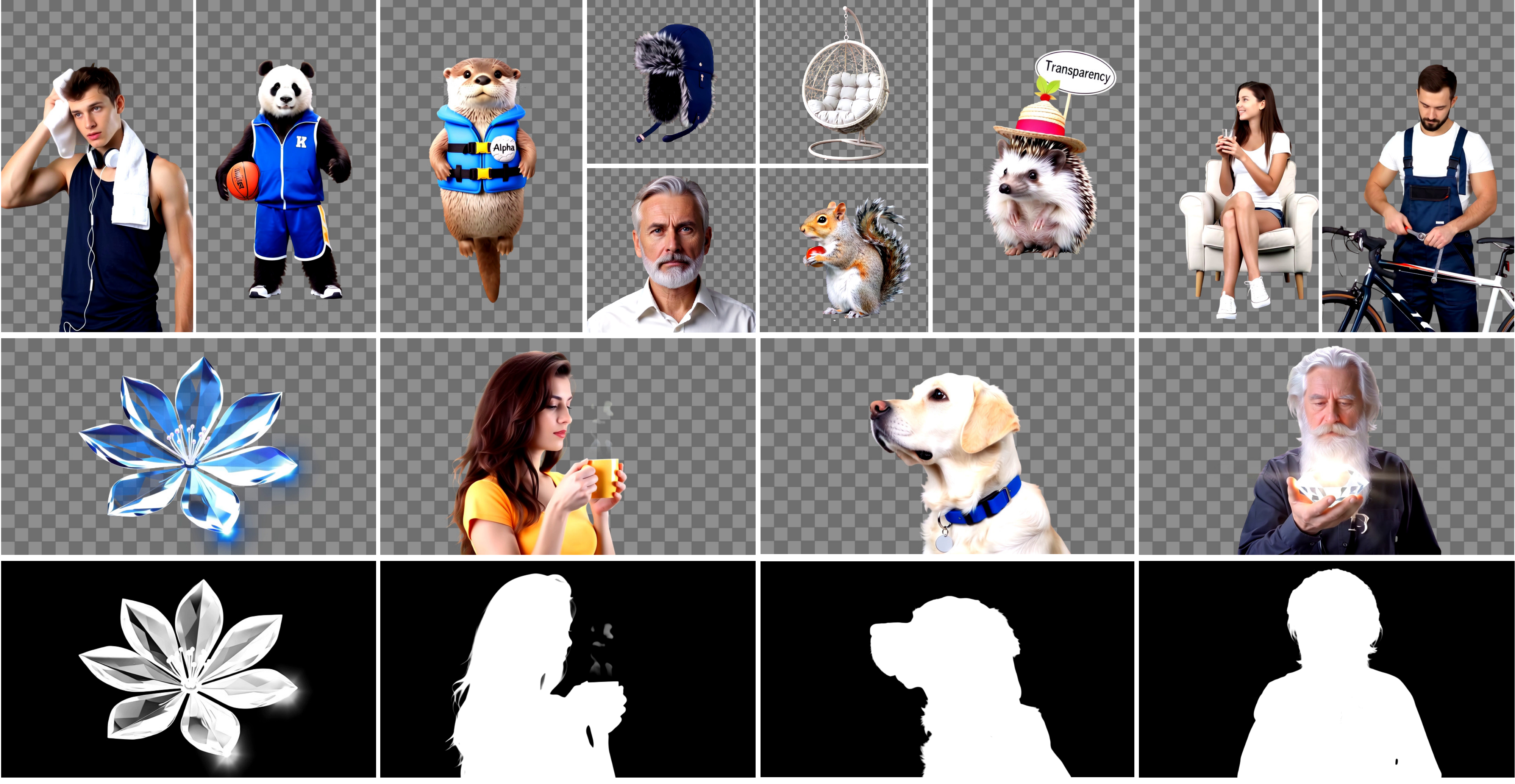}
    \captionof{figure}{Our RGB-A video generation model successfully generates various scenes with accurate and clearly rendered transparency. Notably, it can synthesize diverse semi-transparent objects, and fine-grained details such as hair. See the supplementary material for videos.}\label{fig:teaser}
    \vspace{5mm}
\end{center}%
}]

\begin{abstract}

Generating RGB-A videos, which include alpha channels for transparency, has wide applications. However, current methods often suffer from low quality due to confusion between RGB and alpha. In this paper, we address this problem by learning shiftable RGB‑A distributions. We adjust both the latent space and noise space, shifting the alpha distribution outward while preserving the RGB distribution, thereby enabling stable transparency generation without compromising RGB quality. Specifically, for the latent space, we propose a transparency‑aware bidirectional diffusion loss during VAE training, which shifts the RGB‑A distribution according to likelihood. For the noise space, we propose shifting the mean of diffusion noise sampling and applying a Gaussian ellipse mask to provide transparency guidance and controllability. Additionally, we construct a high‑quality RGB‑A video dataset. Compared to state‑of‑the‑art methods, our model excels in visual quality, naturalness, transparency rendering, inference convenience, and controllability. The released model is available on our website: \href{https://donghaotian123.github.io/Wan-Alpha/}{https://donghaotian123.github.io/Wan-Alpha/}.
\end{abstract}
\section{Introduction}
\label{sec:intro}
%

RGB-A includes an additional alpha channel to represent transparency. RGB-A videos are widely used in creative fields such as game design, filmmaking, television production, and UI development.
Despite its significance, the automatic generation of RGB-A videos remains challenging and has received limited attention.

Early attempts at RGB-A video generation~\cite{chen2025transanimate,TransVDM,layeranimate,ILDiff} adapt image-based methods to the video domain, typically applying the 2D RGB-A VAE from LayerDiffuse~\cite{LayerDiffuse} to frameworks like AnimateDiff~\cite{AnimateDiff}. 
However, this VAE struggles with temporal modeling and entangles RGB and alpha distributions in the latent space, requiring large datasets for adaptation. As a result, these methods produce degraded RGB-A videos with inaccurate transparency and limited motion.
TransPixeler~\citep{wang2025transpixeler} is the state-of-the-art for RGB-A video generation.
It introduces alpha tokens, duplicates the backbone, and uses cross-RGB-A attention.
However, this attention alone fails to effectively capture RGB–alpha relationships, resulting in poor visual quality and unstable transparency.
Moreover, trained on a human matting dataset with mostly opaque subjects, TransPixeler struggles to generalize to transparent cases like veils and smoke.

The core challenge in RGB-A video generation is effectively learning and disentangling the RGB and alpha distributions. To address this, we introduce a novel shiftable RGB-A distribution learner. Unlike previous methods that leave the distributions unaltered, our approach shifts the alpha distribution while preserving the original RGB, enabling a clearer separation that improves visual quality and transparency stability. Moreover, the shift is controllable, allowing users to guide the shape and position of transparency.

We propose to learn shiftable RGB-A distributions in both the noise space and the latent space, which represent the beginning and end of the diffusion process, respectively. This can provide comprehensive guidance throughout the generation.
Regarding how to shift, instead of explicitly increasing the distance between RGB and alpha—which could destabilize training—we introduce smarter and more learnable strategies.
In the \textit{latent space}, we introduce a transparency-aware bidirectional diffusion loss for training the VAE. We employ a frozen video generation model and train the VAE to enhance its diffusion performance in opaque regions while suppressing it in transparent regions. This bidirectional loss implicitly adjusts the distribution and bridges the gap between VAE latent learning and RGB-A video generation. Besides, we employ a series of rendering losses to improve VAE's reconstruction quality.
In the \textit{noise space}, we propose a transparency-guided mean sampler. During video generation training, we create a Gaussian ellipse mask based on transparency and use it to spatially adjust the mean of the diffusion noise. This strategy helps the generation model more effectively distinguish between RGB and alpha. Moreover, during inference, the mask can serve as a guidance tool, enabling users to control transparency.

Furthermore, to address the scarcity of high-quality RGB-A videos, we carefully select and process data to ensure high resolution, smooth motion, rich content, detailed captions, and diverse attribute labels.
Combining our well-designed framework with the high-quality dataset, our model achieves state-of-the-art performance across multiple metrics.
As shown in Fig.~\ref{fig:teaser}, our model successfully generates various scenes with accurate and clearly rendered transparency.
Our core contributions are as follows:


\begin{itemize}
    \item We develop a novel model for high-quality RGB-A video generation. The key idea is to learn shiftable RGB-A distributions, enabling stable and controllable alpha generation without compromising RGB quality.
    \item We propose a bidirectional diffusion loss for distinguishable RGB-A latent learning and a mean-shifting strategy for controllable RGB-A generation. These designs effectively introduce transparency guidance and promote joint RGB-A distribution learning.
    \item Along with our well-designed framework, we build a high-quality dataset. All components work together, enabling our model to outperform existing methods across multiple metrics and achieve flexible controllability.
\end{itemize}

Our model and dataset will be open-source and freely accessible to the public.

\section{Related Work}
\label{sec:related}


\noindent {\bf RGB-A Image Generation~}~Compared to the common RGB modality, RGB-A data is significantly scarcer and more difficult to acquire. This limited data availability poses significant challenges for training large-scale RGB-A generation models from scratch. To address the challenge, researchers leverage pretrained RGB generation models.
LayerDiffuse~\cite{LayerDiffuse} is the first method for RGB-A image generation.
It adapts Stable Diffusion~\cite{latentdiffusion} by training its VAE to embed transparency into the RGB latent space, followed by fine-tuning using LoRA.
AlphaVAE~\cite{alphavae} improves the RGB-A VAE and reduces the training images from 1M to just 8K samples. Alfie~\cite{quattrini2024alfie} proposes a training-free method that infers alpha values directly from attention maps. However, it can only generate images with transparent backgrounds, which neglects the semi-transparent effects. Some research extends transparent generation to multi-layer generation~\cite{DreamLayer,yang2025generative,huang2025psdiffusion,kang2025layeringdiff} and transparent image inpainting~\cite{dai2025trans}.
Compared to RGB-A images, RGB-A videos offer greater visual richness, but creating them presents a more significant challenge.


\vspace{1mm}
\noindent {\bf RGB-A Video Generation~}~RGB-A video generation presents a greater challenge than RGB-A image generation due to the need for joint learning of the temporal dimension. Additionally, high-quality RGB-A video datasets remain extremely scarce.
To tackle these challenges, several approaches~\cite{chen2025transanimate,TransVDM,layeranimate,ILDiff} have combined the LayerDiffuse VAE with video generation models. However, the LayerDiffuse VAE, originally designed for images, does not adequately address temporal consistency in video sequences.
LayerFlow handles multi-layer generation~\cite{ji2025layerflow}.
TransPixeler~\cite{wang2025transpixeler}, the current state-of-the-art model for transparent video generation, duplicates the backbone network to create an alpha version and employs cross-RGB-A attention for RGB-alpha information exchange. However, this duplication results in a significant increase in inference cost. Additionally, existing methods fail to effectively distinguish between RGB and alpha channels, leading to unsatisfactory quality and unstable transparency.
In this work, we propose a novel RGB-A distribution learner that facilitates the learning of RGB-A video generation, resulting in improved generation performance.

\section{Method}
\label{sec:method}


To develop a high-quality RGB-A video generation model, we first train an RGB-A VAE with transparency-aware bidirectional diffusion loss in Sec.~\ref{sec:vae}. Then, we train the video generation with transparency-guided mean sampler in Sec.~\ref{sec:t2v}. Besides, we build a high-quality dataset in Sec.~\ref{sec:dataset}.

\begin{figure*}[t]
    \centering
    \includegraphics[width=0.99\linewidth]{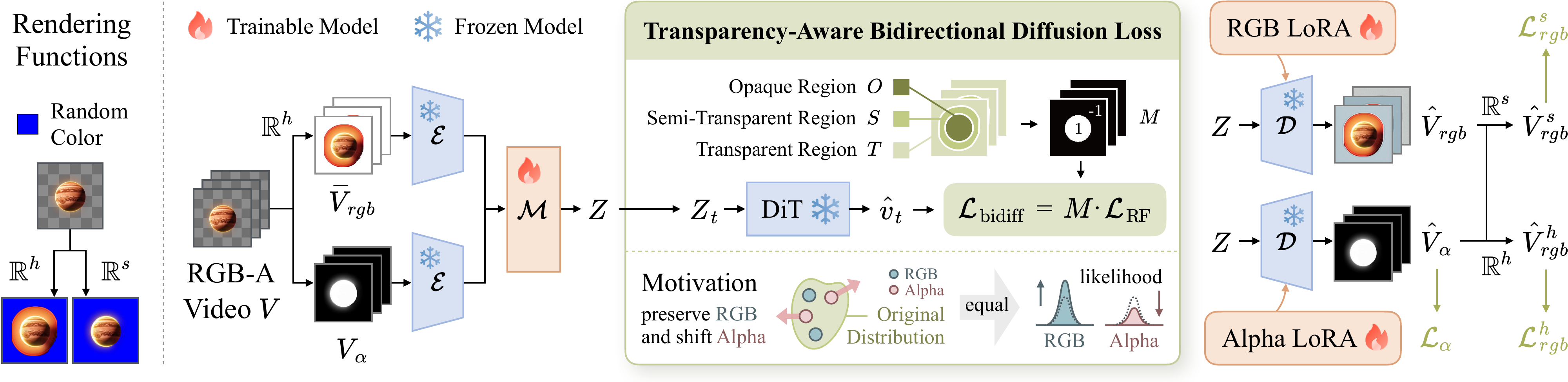}
    \vspace{-2mm}
    \caption{The training framework of our VAE. The RGB-A video $V$ is split into the RGB video $V_{rgb}$ and the alpha video $V_{\alpha}$. $V_{rgb}$ is hard-rendered into $\bar{V}_{rgb}$. $\bar{V}_{rgb}$ and $V_{\alpha}$ are then fed into the frozen VAE encoder $\mathcal{E}$, producing two distinct feature representations. Next, a feature merging block $\mathcal{M}$ combines these features to generate the latent feature $Z$. Finally, $Z$ is passed into both the RGB and alpha VAE decoders $\mathcal{D}$, which predict the RGB video $\hat{V}_{rgb}$ and the alpha video $\hat{V}_{\alpha}$. To guide VAE training, we adopt a transparency-aware bidirectional diffusion loss $\mathcal{L}_\mathrm{bidiff}$, and a series of reconstruction losses $\mathcal{L}_\alpha$, $\mathcal{L}^s_{rgb}$, and $\mathcal{L}^h_{rgb}$. }
    \label{fig:vae}
\end{figure*}

\subsection{Distinguishable RGB-A Latent Learning}
\label{sec:vae}



We start from a pretrained RGB video generation model~\cite{wan2025wan}, which uses a VAE to encode videos into latent representations and a diffusion transformer (DiT) to map noise to these latents.
Following LayerDiffuse~\cite{LayerDiffuse}, we extend the VAE to embed RGB and alpha channels into a shared latent space, keeping the latent dimensions unchanged. This allows us to fine-tune the DiT for RGB-A video generation without any architectural modification.
However, directly encoding RGB and alpha together without additional guidance often leads to an entangled latent space.
As a result, the DiT may predict incorrect transparency, producing holes and artifacts in the generated RGB-A videos.

\vspace{1mm}
\noindent{\bf Transparency-Aware Bidirectional Diffusion Loss~}~We propose to improve the latent space by preserving the RGB distribution while shifting the alpha distribution. The preserved RGB distribution remains familiar to the DiT, maintaining the generation capability of the base model. The shifted alpha distribution enhances the separation between opaque and transparent regions, leading to better transparency synthesis.

When attempting to shift the alpha distribution, one key challenge arises: there is a gap between the latent-space distance and the generation capability of DiT. Statistically separating RGB and alpha latents does not necessarily ensure that DiT can better distinguish between them and may even degrade its generation ability.
To address this issue, we introduce a frozen DiT into the VAE training process. From the perspective of DiT, preserving the RGB distribution while shifting the alpha distribution away is equivalent to increasing the likelihood of RGB and decreasing the likelihood of alpha. This can be implemented via a bidirectional diffusion loss, where a mask reverses the diffusion loss for transparent regions. In this manner, the VAE can learn RGB-A latents that are more distinguishable to DiT.

Specifically, we resize the alpha video to match the latent shape, and define the mask $M$ as follows:
\begin{equation}
\begin{aligned}
    M(p) = \begin{cases}
        \phantom{-}1, & \text{if } p \in O, \\
        -1, & \text{if } p \in S \cup T,
    \end{cases}
\end{aligned}
\end{equation}
where $p$ represents a pixel in the latent feature space, and $O$, $S$, and $T$ denote the opaque, semi-transparent, and transparent regions, respectively.

Then, we employ $M$ to invert the loss sign within the non‑opaque regions.
The original Rectified Flow~\cite{lipman2023flow,esser2024scaling} (RF) training objective is formulated as:
\begin{align}
    Z_{t} \!=\! t \!\cdot\! \epsilon \!+\! (1 - t)\! \cdot\! Z,  
    v_{t} = \epsilon - Z, 
    \mathcal{L}_{\mathrm{RF}} = \left\| \hat{v}_{t} - v_t \right\|^2, \label{eq:t2v}
\end{align}
where $Z$ denotes the encoded latent, $t$ denotes the timestep, 
$\epsilon$ denotes the noise, and $\hat{v}_{t}$ denotes the output predicted by the DiT model. 
Our final transparency-aware bidirectional diffusion loss $\mathcal{L}_{\mathrm{bidiff}}$ is defined as:
\begin{equation}
    \mathcal{L}_{\mathrm{bidiff}} = M \cdot \mathcal{L}_{\mathrm{RF}}.
\end{equation}

\begin{figure*}[t]
    \centering
    \includegraphics[width=0.99\linewidth]{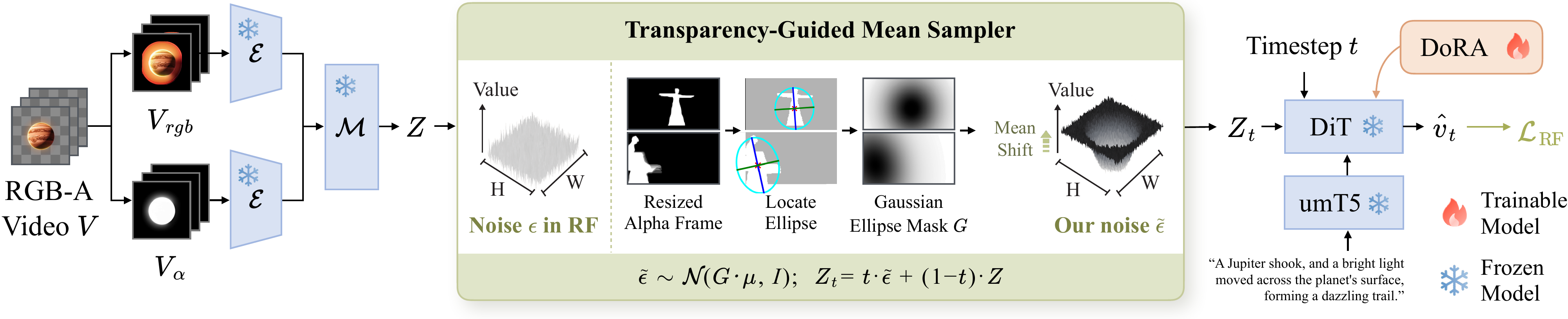}
    \vspace{-2mm}
    \caption{Our RGB–A video generation training framework. RGB-A videos are first encoded into latents using our VAE, after which a DiT is trained on these latents with DoRA.  To incorporate transparency guidance, we employ a mean sampler. Specifically, we predict an ellipse from the alpha frame that approximately covers the opaque content. This ellipse is then converted into a Gaussian mask $G$, which is used to shift the mean of the noise within transparent regions.  Finally, the noise term \(\tilde{\epsilon} \sim \mathcal{N}(G, 1)\) is applied during both training and inference.}
    \label{fig:t2v}
\end{figure*}

\noindent{\bf Rendering Losses~}~We design a set of losses to enhance reconstruction quality. First, we define soft render $\mathbb{R}^s$ and hard render $\mathbb{R}^h$, as follows:
\begin{align}
    \mathbb{R}^s (V_{rgb}, V_\alpha, c) &= V_{rgb} \cdot V_\alpha + c \cdot (1 - V_\alpha), \\
    \mathbb{R}^h (V_{rgb}, V_\alpha, c) &= V_{rgb} \cdot \mathbf{1}_{V_\alpha > 0} + c \cdot ( 1 - \mathbf{1}_{V_\alpha > 0} ),
\end{align}
where $c$ is a random color from a predefined color set $C=\{$black, blue, green, cyan, red, magenta, yellow, white$\}$. $\mathbf{1}_{\alpha > 0}$ denotes the indicator function, which equals 1 when $\alpha > 0$ and 0 otherwise.

To guide the model to focus differently on opaque, semi-transparent, and transparent regions, we apply the reconstruction loss to three modalities: the alpha video $\hat{V}_\alpha$, the soft-rendered video $\hat{V}^s_{rgb}$, and the hard-rendered video $\hat{V}^h_{rgb}$. The latter two are formulated as follows:
\begin{equation}
\hat{V}^s_{rgb} = \mathbb{R}^{s}(\hat{V}_{rgb}, \hat{V}_{\alpha}, c_s), \quad
\hat{V}^h_{rgb} = \mathbb{R}^{h}(\hat{V}_{rgb}, \hat{V}_{\alpha}, c_h),
\end{equation}
where $c_s$ and $c_h$ represent random colors sampled from $C$. $\hat{V}_{rgb}$ denotes the reconstructed RGB video, and $\hat{V}_\alpha$ denotes the reconstructed alpha channel.

For each modality, we employ a composite reconstruction objective that combines pixel-wise, perceptual~\cite{perceptual_loss}, and edge-consistency terms. Given a predicted video $\hat{V}$ and its corresponding ground-truth $V$, the overall loss is defined as:
\begin{equation}
\begin{split}
&\mathcal{L}_\Phi = ||\Phi(\hat{V})-\Phi({V})||_2,~ \mathcal{L}_s = ||S(\hat{V})-S({V})||, \\
&~~~~~~~~~\mathcal{L}_{recon}(\hat{V}, {V})= ||\hat{V}-{V}|| + \mathcal{L}_\Phi + \mathcal{L}_s,
\end{split}
\end{equation}
where $\Phi(\cdot)$ denotes the VGG feature extractor~\cite{vgg}, and $S(\cdot)$ denotes the Sobel operator~\cite{Sobel1990AnI3}.
We apply this composite loss to the alpha channel, the soft-rendered RGB video, and the hard-rendered RGB video:
\begin{equation}
\begin{aligned}
    &\mathcal{L}_{\alpha} \!=\! \mathcal{L}_{recon}(\hat{V}_{\!\alpha},\! {V}_{\!\alpha}), \ \ \ \mathcal{L}_{rgb}^s \!=\! \mathcal{L} _{recon}(\hat{V}_{\!rgb}^s,\! {V}_{\!rgb}^s),\\ &\mathcal{L}_{rgb}^h \!=\! \mathcal{L}_{recon}(\hat{V}_{\!rgb}^h, \!{V}_{\!rgb}^h).
\end{aligned}
\end{equation}
The final training objective for VAE is defined as follows:
\begin{equation} \mathcal{L}_{vae}=\mathcal{L}_{\alpha}+\mathcal{L}^s_{rgb}+\mathcal{L}^h_{rgb}+\mathcal{L}_\mathrm{bidiff}.
\end{equation}

\noindent{\bf Architecture~}~The overall framework is illustrated in Fig.~\ref{fig:vae}. First, we split the RGB-A video into an RGB video and an alpha video, where we duplicate the alpha channel three times. To prevent the encoder from confusing the concepts of RGB background color with transparency, we randomly select a color $\bar{c}$ from the color set $C$ and apply the hard render function $\mathbb{R}^h$ to the RGB video $V_{rgb}$, resulting in a rendered video $\bar{V}_{rgb} = \mathbb{R}^{h}({V}_{rgb}, {V}_{\alpha}, \bar{c})$.
We feed $\bar{V}_{rgb}$ and $V_{\alpha}$ into the frozen VAE encoder $\mathcal{E}$. Then, we use the feature merge block $\mathcal{M}$ to fuse the RGB and alpha features, which is constructed from a series of causal residual blocks~\cite{wan2025wan} and attention layers. The predicted latent $Z$ is given by:
\begin{equation}
    Z=\mathcal{M}(\mathcal{E}(\bar{V}_{rgb}),\mathcal{E}(V_{\alpha})).
\end{equation}
Next, the latent $Z$ is fed into the frozen VAE decoders, which are equipped with RGB LoRA~\cite{hu2022lora} and alpha LoRA, respectively.
The reconstructed RGB video $\hat{V}_{rgb}$ and alpha video $\hat{V}_{\alpha}$ are formulated as:
\begin{equation}
    \hat{V}_{rgb} = \mathcal{D}_{\text{w/ RGB LoRA}}(Z),\\
    ~~~\hat{V}_{\alpha} = \mathcal{D}_{\text{w/ Alpha LoRA}}(Z).
\end{equation}


\subsection{Controllable RGB-A Generation}
\label{sec:t2v}

In RGB video generation, even without specifying a background, models typically synthesize background elements to enrich visual content and enhance realism. In contrast, for RGB-A video generation, users often prefer a clean, fully transparent background. When adapting a pre-trained RGB model to RGB-A, preserving the generative capabilities of the base model can improve performance; however, it also tends to introduce unwanted backgrounds. Existing approaches either ignore this issue or require overfitting to RGB-A data, which causes the model to lose the generative abilities of the base model and results in degraded performance.
To address this issue, we propose incorporating transparency priors into the Gaussian noise. This encourages the video generation model to better distinguish between opaque and transparent regions, while also providing a controllable mechanism for transparency.

\vspace{1mm}
\noindent{\bf Transparency-Guided Mean Sampler~}~We propose to shift the mean of the Gaussian noise according to the input alpha video in the rectified flow framework.
Similar to \cite{BBDM,Shifted_Diffusion,ShiftDDPMs,TKG-DM}, let $\mu(Z)$ be a mean function determined by the input latent variable $Z$. 
We define the modified noise as follows:
\begin{equation}
    \tilde{\epsilon} \sim \mathcal{N}\big(\mu(Z), I\big).
\end{equation}
The interpolated latent $Z_t$ at time $t$ and the corresponding velocity field $\tilde{v}_t$ are formulated as follows:
\begin{equation}
Z_t = t\, \tilde{\epsilon} + (1-t)\, Z,~~\tilde{v}_t = Z_t - \tilde{\epsilon}
\end{equation}
Our modified training objective for rectified flow becomes:
\begin{equation}
    \mathcal{L}_{\mathrm{RF}} = \left\| \hat{v}_{t} - \tilde{v}_t \right\|^2,
\end{equation}
where $\hat{v}_{t}$ denotes the output of DiT. During inference, we start from $\mathcal{N}(\mu, I)$, where the mean $\mu$ can be user-defined.

For designing the $\mu(\cdot)$ function, we propose generating a Gaussian ellipse mask based on the alpha channel. This essentially conveys the approximate shape and position of transparent regions, while still allowing the model sufficient flexibility to determine the finer structural details and motion.
Specifically, we first resize alpha frames to match the resolution of the latents, and binarize each resized alpha frame $A \in \mathbb{R}^{H \times W}$ to obtain a hard alpha mask $B = \mathbb{I}(A > 0.5)$, which defines a point set:
\begin{equation}
\mathcal{P} = \{(x, y) \mid B(x, y) = 1\}.
\end{equation}
We compute the mean position $\boldsymbol{\mu}\;=\;(\mu_x, \mu_y)^\top$ and the covariance matrix of the point set:
\begin{equation}
\boldsymbol{\Sigma}
= \frac{1}{|\mathcal{P}|-1}\sum_{(x,y)\in\mathcal{P}}
\big[(x, y)^\top - \boldsymbol{\mu}\big]
\big[(x, y)^\top - \boldsymbol{\mu}\big]^\top.
\end{equation}
Then we perform eigenvalue decomposition $\boldsymbol{\Sigma} = \mathbf{V}\boldsymbol{\Lambda}\mathbf{V}^\top$ to obtain eigenvectors $\mathbf{v}_1, \mathbf{v}_2\in \mathbf{V}$
and eigenvalues $\lambda_1, \lambda_2 \in \boldsymbol{\Lambda}$, which correspond to the major and minor directions of the distribution. Each point $(x, y) \in \mathcal{P}$ is projected onto the principal axes, and the difference between the maximum and minimum projections determines the axis lengths $(a, b)$ of the fitted ellipse. The orientation angle is computed as:
\begin{equation}
\theta = \arctan2(v_{1y}, v_{1x}),
\end{equation}
where $(v_{1x}, v_{1y})$ are the components of the principal direction $\mathbf{v}_1$. Finally, we construct a Gaussian ellipse mask $G$ aligned with the estimated geometry:
\begin{equation}
G(x, y) = \exp\!\left(
-\frac{1}{2}
\left[
\left(\frac{x'}{a/2}\right)^2 +
\left(\frac{y'}{b/2}\right)^2
\right]
\right),
\end{equation}
where $x'$, $y'$ is the rotated $x,y$ aligned with the fitted ellipse. The resulting $G$ is normalized to $[0,1]$ to produce the final Gaussian ellipse mask. We further introduce a factor $\mu$ to control the strength of the mean shift:
\begin{equation}
    \tilde{\epsilon} \sim \mathcal{N}(G \cdot \mu,I).
\end{equation}


\vspace{1mm}
\noindent{\bf Architecture~}~As illustrated in Fig.~\ref{fig:t2v}, we employ our pretrained RGB-A VAE encoder to obtain the latent $Z$. 
The proposed noise term \(\tilde{\epsilon}\) is then applied to produce the noisy latent \(Z_t\). 
The text condition is encoded using the umT5 text encoder~\cite{chung2023unimax}. 
The diffusion transformer is trained with DoRA~\cite{liu2024dora}, which we find yields better semantic alignment and higher-quality video generation compared to LoRA.


\vspace{1mm}
\noindent{\bf Inference~}~Our framework is straightforward to deploy. Compared to the base model~\cite{wan2025wan}, our RGB-A model only requires modifying the initial noise, duplicating the VAE decoder, and loading the RGB and alpha decoder LoRAs along with the DiT DoRA. Notably, the LoRA and DoRA parameters can be fully merged into the base model, introducing no extra computational cost. By default, $G$ is placed at the center.
Furthermore, since we largely preserve the inference architecture and maintain the original RGB distribution, we are able to leverage the acceleration tools of the base model. In particular, we adopt LightX2V~\cite{lightx2v} to speed up video generation, achieving high-quality results in only 4 sampling steps without the need for classifier-free guidance~\cite{CFG}.
The state-of-the-art TransPixeler~\cite{wang2025transpixeler} requires 32 minutes to generate 49 frames, $480\times720$, 8 FPS. In contrast, our model generates 81 frames at a resolution of $480\times832$ and 16 FPS in just 128 seconds, which is 15 times faster.

\begin{figure}
    \centering
    \includegraphics[width=0.99\linewidth]{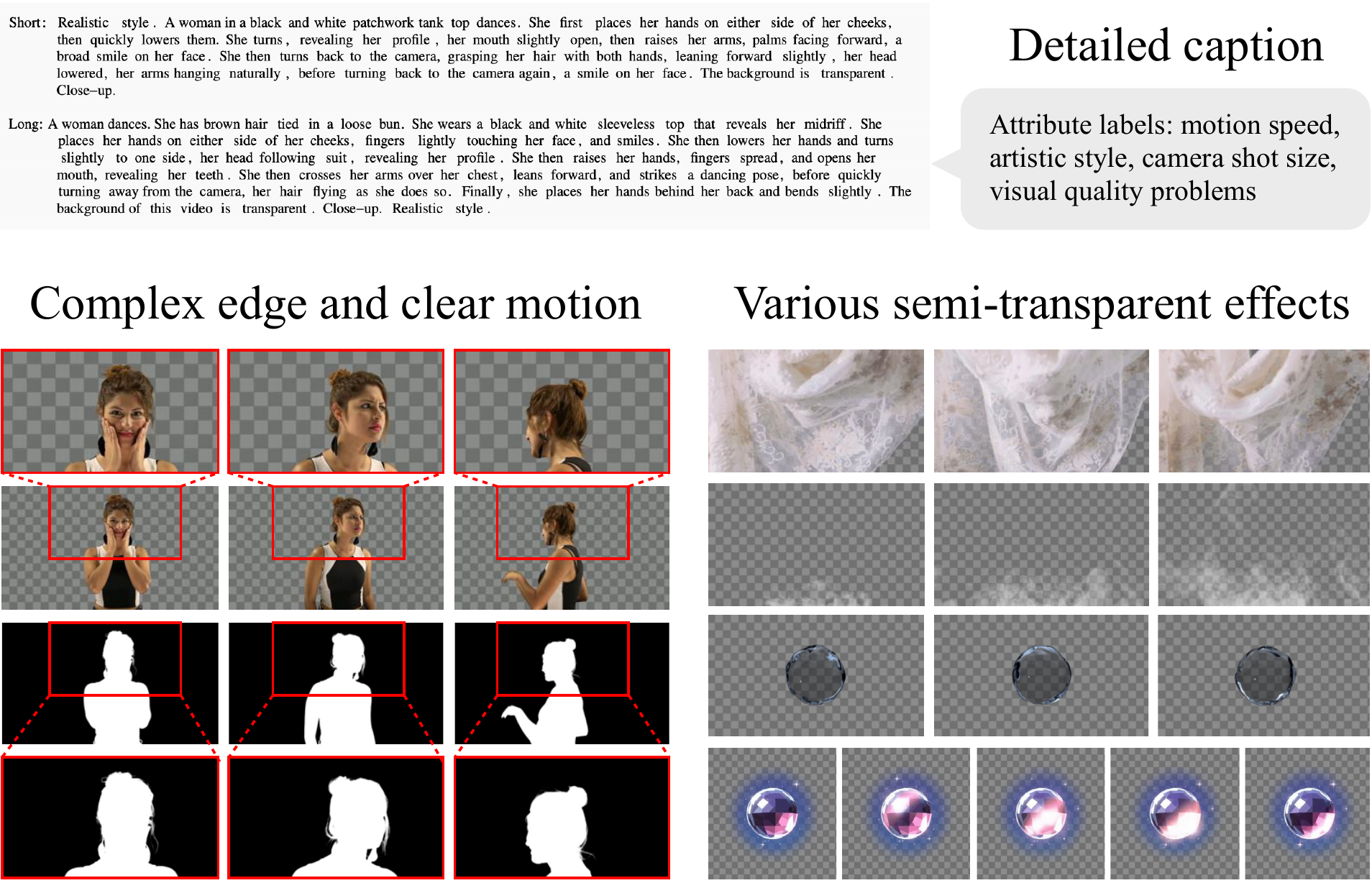}
    \caption{Samples from our dataset. Our videos show clear motion and complex edges, such as hair. Each video has short and long captions, plus attribute labels. The dataset also includes various semi‑transparent effects like sheer fabric, smoke, water, and glow.}
    \label{fig:dataset}
\end{figure}

\begin{figure*}
    \centering
    \includegraphics[width=0.99\linewidth]{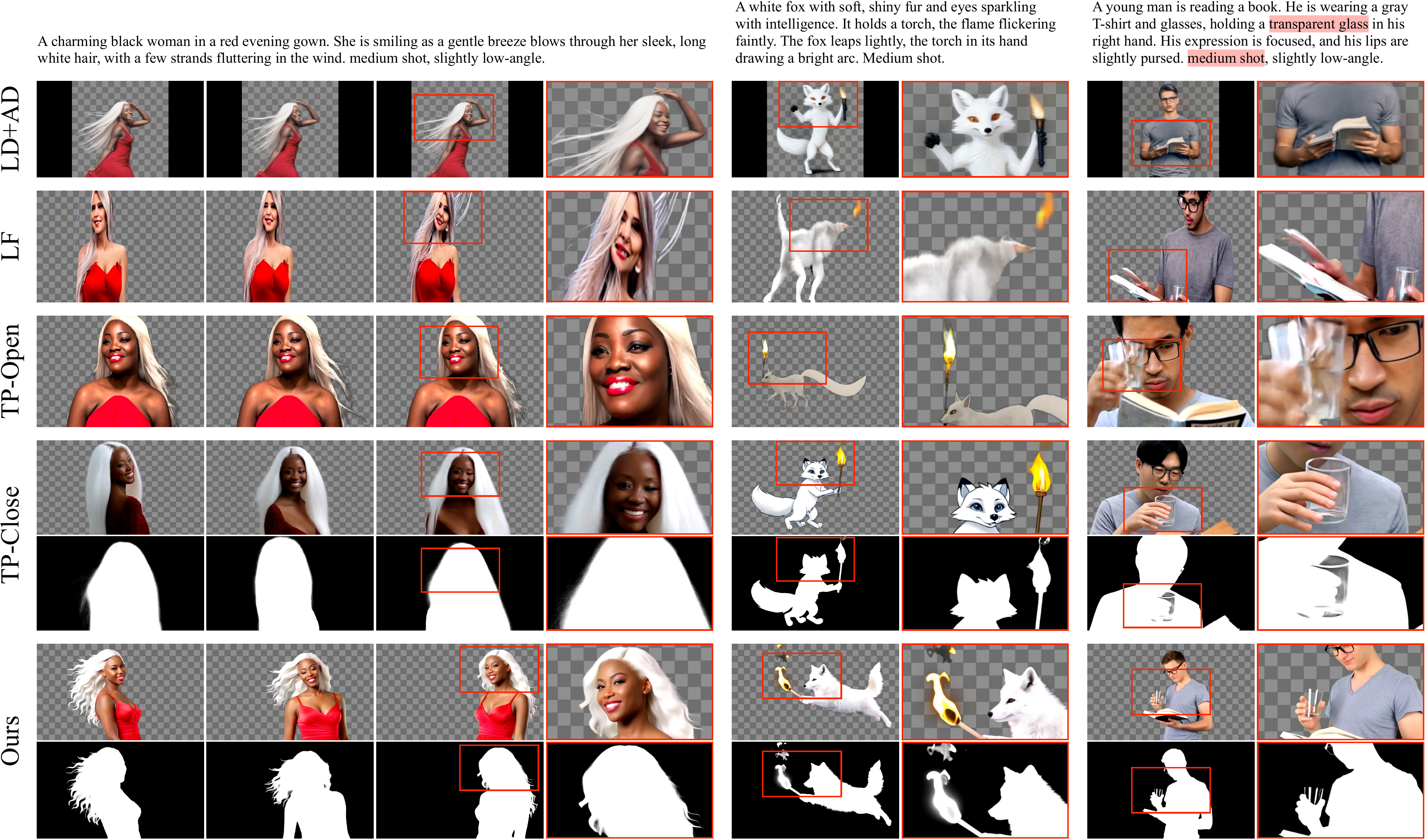}
    \caption{Comparison with existing RGB-A video generation methods. From top to bottom: LayerDiffuse + AnimateDiff (LD+AD), LayerFlow (Single) [LF], TransPixeler (Open) [TP-Open], TransPixeler (Close) [TP-Close], and our model. Our model shows superior aesthetics, realism, motion consistency, and transparency accuracy. Please refer to the supplementary material for a video demonstration.}
    \label{fig:comparison}
\end{figure*}

\begin{figure*}
    \centering
    \includegraphics[width=0.99\linewidth]{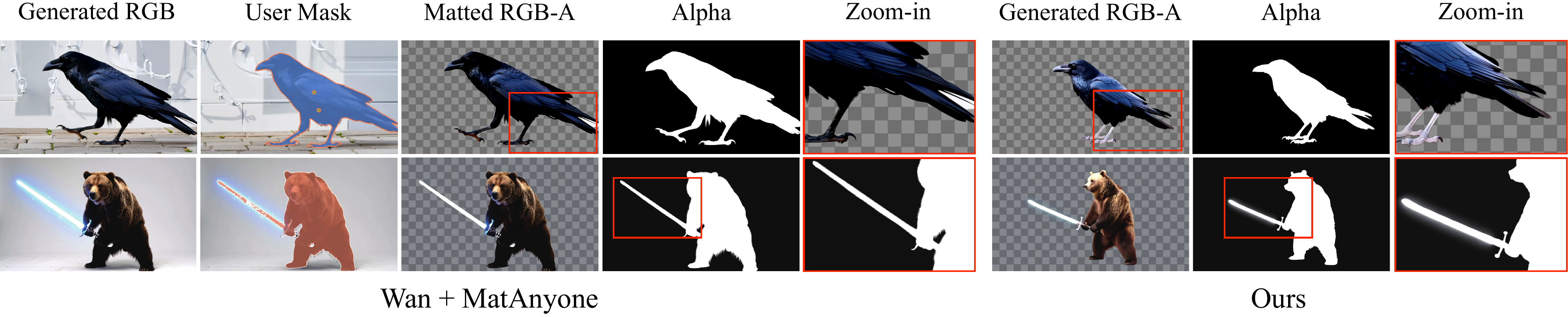}
    \vspace{-2mm}
    \caption{In comparison with the video generation plus video matting solution, our model demonstrates higher transparency quality.}
    \label{fig:mat}
\end{figure*}

\subsection{Dataset}
\label{sec:dataset}

Our VAE training data consists of a diverse set of RGB-A images and videos, collected from 10 image matting datasets and 3 video matting datasets.
Images are collected from AIM-500~\cite{AIM}, AM-2K~\cite{li2022bridging}, Distinctions-646~\cite{Qiao_2020_CVPR}, HHM2K~\cite{Sun_2023_CVPR}, HHM50K~\cite{Sun_2023_CVPR}, Human2K~\cite{Liu_2021_ICCV}, P3M-10K~\cite{rethink_p3m}, RealWorldPortrait-636~\cite{yu2021mask}, SIMD~\cite{sun2021sim}, and Transparent-460~\cite{cai2022TransMatting}.
We convert images into static videos and then randomly slide a window along the temporal axis to simulate motion.
Videos are collected from DVM~\cite{sun2021dvm}, VideoMatting108~\cite{zhang2021attention}, VideoMatte240K~\cite{lin2021real}, as well as additional sources gathered from the Internet. The dataset is split into 95\% for training and 5\% for validation. This split results in 77,237 training videos and 4,066 validation videos.

Our RGB-A video generation data is sourced from the same collection used for VAE training. To achieve high quality, we make careful selections, with a specific focus on examples exhibiting clear motion, semi-transparent objects, and lighting effects. Then, we use the Qwen2.5-VL-72B model~\cite{qwen2.5-VL} to generate initial short and long captions for the 429 samples, and apply a set of specific labels for attributes like motion speed, artistic style, shot size, and visual quality problems. Our final RGB-A video generation dataset includes 301 videos from video matting datasets, 43 images from image matting datasets, and 85 videos with special effects from the Internet. Samples can be found in Fig.~\ref{fig:dataset}.
\section{Experiments}
\label{sec:experiments}
\subsection{Implementation Details}
Our model is built on the pre‑trained Wan2.1-T2V-14B~\cite{wan2025wan}. We generate RGB‑A videos at a resolution of $480\times832$, comprising 81 frames at 16 FPS with only 4 sampling steps. The LoRA rank for the RGB and alpha VAE decoders is 128, and the DoRA rank for DiT is 32. The VAE was trained for 75k iterations with a batch size of 2, while the DiT was trained for 1,750 iterations with a batch size of 8. 
Please refer to the supplementary material for more details.


We use VBench~\cite{huang2023vbench} to assess aesthetics, motion smoothness, and temporal consistency.
Following OpenS2V~\cite{OpenS2V}, we use the VLLM model GPT‑4o~\cite{gpt_4o} to measure text alignment and naturalness.
We render RGB-A videos into white background.
As current evaluation metrics do not support RGB‑A and thus cannot assess transparency, we additionally conducted a user study to evaluate transparency correctness together with a subjective assessment of overall quality. We ask the users to rank all methods and report the average rank.

\setlength{\tabcolsep}{3.4pt}
\renewcommand{\arraystretch}{1.2}
\begin{table*}[t]
\tiny
    \centering
    \caption{Quantitative comparison with existing RGB‑A video generation methods; our model achieves the highest scores across all metrics, not only demonstrating superior overall performance but also validating the effectiveness of the proposed framework design.}
    \vspace{-2mm}
    \footnotesize
    \begin{tabular}{c||ccccc}
    \hline
    \multicolumn{1}{c||}{\textbf{Method}}    
    & \multirow{1}{*}{\bf {Text Alignment$\uparrow$}}
    & \multirow{1}{*}{\bf {Aesthetic Quality$\uparrow$}}
    & \multirow{1}{*}{\bf {Naturalness$\uparrow$}}
    & \multirow{1}{*}{\bf {Motion Smoothness$\uparrow$}}
    & \multirow{1}{*}{\bf {Temporal Flickering$\uparrow$}}\\ \hline\hline
      
     \textbf{LayerFlow (Single)}~\cite{ji2025layerflow}&2.67&0.535&2.35&0.9837 &0.9788\\ \hline\textbf{LayerDiffuse}~\cite{LayerDiffuse}\textbf{ + AnimateDiff}~\cite{AnimateDiff}&3.15&0.617&3.03&0.9893 &0.9853\\ \hline
     
     \textbf{TransPixeler (Open)~\cite{wang2025transpixeler}}&3.16&0.570&2.97& 0.9821 & 0.9872\\ \hline
     \textbf{TransPixeler (Close)~\cite{wang2025transpixeler}}&3.45&0.573&3.07& 0.9907 & 0.9822\\ \hline
     \textbf{Ours}&\bf{4.00}&\bf{0.649}&\bf{3.19}&\bf{0.9949}&\bf{0.9941} \\ \hline
    \end{tabular}
    \label{tab:compare}
\end{table*}

\setlength{\tabcolsep}{5pt}
\renewcommand{\arraystretch}{1.2}
\begin{table}[t]
\tiny
    \centering
    \caption{Users' ranking of transparency correctness and overall quality. Our model is mostly preferred by the users.}
    \vspace{-2mm}
    \footnotesize
    \begin{tabular}{c||ccccc}
    \hline
    \multicolumn{1}{c||}{\textbf{Method}}    
    & \multirow{1}{*}{\bf {Transparency$\downarrow$}}
    & \multirow{1}{*}{\bf {Overall$\downarrow$}} \\
    \hline\hline
     \textbf{LayerFlow (Single)}~\cite{ji2025layerflow} & 4.29 &3.57 \\ \hline
     \textbf{LayerDiffuse}~\cite{LayerDiffuse}\textbf{ + AnimateDiff}~\cite{AnimateDiff} & 3.40 &4.23 \\ \hline
     \textbf{TransPixeler (Open)}~\cite{wang2025transpixeler} & 2.51 &2.71  \\ \hline
     \textbf{TransPixeler (Close)}~\cite{wang2025transpixeler} & 2.57 &3.37 \\ \hline
     \textbf{Ours} & \textbf{1.23} &\textbf{1.11} \\ \hline
    \end{tabular}
    \label{tab:userstudy}
\end{table}

\subsection{Comparison Results}
Since RGB-A video generation is a relatively new task, only TransPixeler~\cite{wang2025transpixeler} and LayerFlow~\cite{ji2025layerflow} are currently capable of producing RGB-A videos. 
For TransPixeler, in addition to its open-source version, we also evaluate its closed-source version on Adobe Firefly. 
LayerFlow is a multi-layer video generation framework, and we focus our evaluation on its foreground-layer generation capability. 
To enable a more comprehensive comparison, we further implement an additional baseline that integrates LayerDiffuse~\cite{LayerDiffuse} into the text-to-video model AnimateDiff~\cite{AnimateDiff}.

Quantitative results are shown in Table~\ref{tab:compare}, and qualitative results are provided in Fig.~\ref{fig:comparison}. The combination of LayerDiffuse and AnimateDiff exhibits poor motion, unnatural visual content, and inadequate text alignment (e.g., failing to generate the glass in the third case of Fig.~\ref{fig:comparison}). Correspondingly, in Table~\ref{tab:compare}, LayerDiffuse+AnimateDiff attains low performance across all quantitative metrics. These results suggest that merely adding RGB-A image LoRAs and VAEs to a video generation model is inadequate for achieving high-quality RGB-A video generation.
LayerFlow and TransPixeler are both extended from a pretrained RGB video generation model. LayerFlow produces distorted objects, while TransPixeler yields better results but still lacks realism and naturalness in fantasy scenes (e.g., a fox holding a torch, the second case of Fig.~\ref{fig:comparison}) and exhibits incorrect visual content (e.g., wrong fingers, the third case of Fig.~\ref{fig:comparison}). This is because these methods degrade the RGB generation capability. Furthermore, the open version of TransPixeler generates incorrect glass transparency, whereas the closed version wrongly produces a transparent glass when it should be opaque, as it is positioned in front of a person. These issues stem from the models’ inability to effectively distinguish between RGB and alpha channels, leading to wrong transparency.
Benefiting from our framework design, which preserves the RGB distribution to maintain the base model’s capabilities and shifts the alpha distribution to better distinguish transparency, our model achieves the highest scores across all quantitative evaluation metrics in Table~\ref{tab:compare}, and is able to generate vivid hair edges, natural human motion, realistic fire and smoke effects, as well as transparent glass and glasses, as shown in Fig.~\ref{fig:comparison}.
User study results in Table~\ref{tab:userstudy} also demonstrate our superior performance.

We also present a comparison against combining a video generation model with video matting. Specifically, we use our base model Wan~\cite{wan2025wan} in conjunction with the state-of-the-art video matting method MatAnyone~\cite{MatAnyone}. It is important to note that MatAnyone requires the user to provide a mask, and thus the Wan+MatAnyone solution is not fully end-to-end automatic. To reduce the matting burden, we instruct Wan to generate a white background. 
However, the matting results remain unsatisfactory: as shown in Fig.~\ref{fig:mat}, portions of the white background are still not removed, and Wan+MatAnyone is incapable of handling semi-transparent effects such as lighting. In contrast, our model is able to generate accurate transparency.



\subsection{Ablation Study}
\begin{figure}[t]
    \centering
    \vspace{-1mm}
    \includegraphics[width=0.99\linewidth]{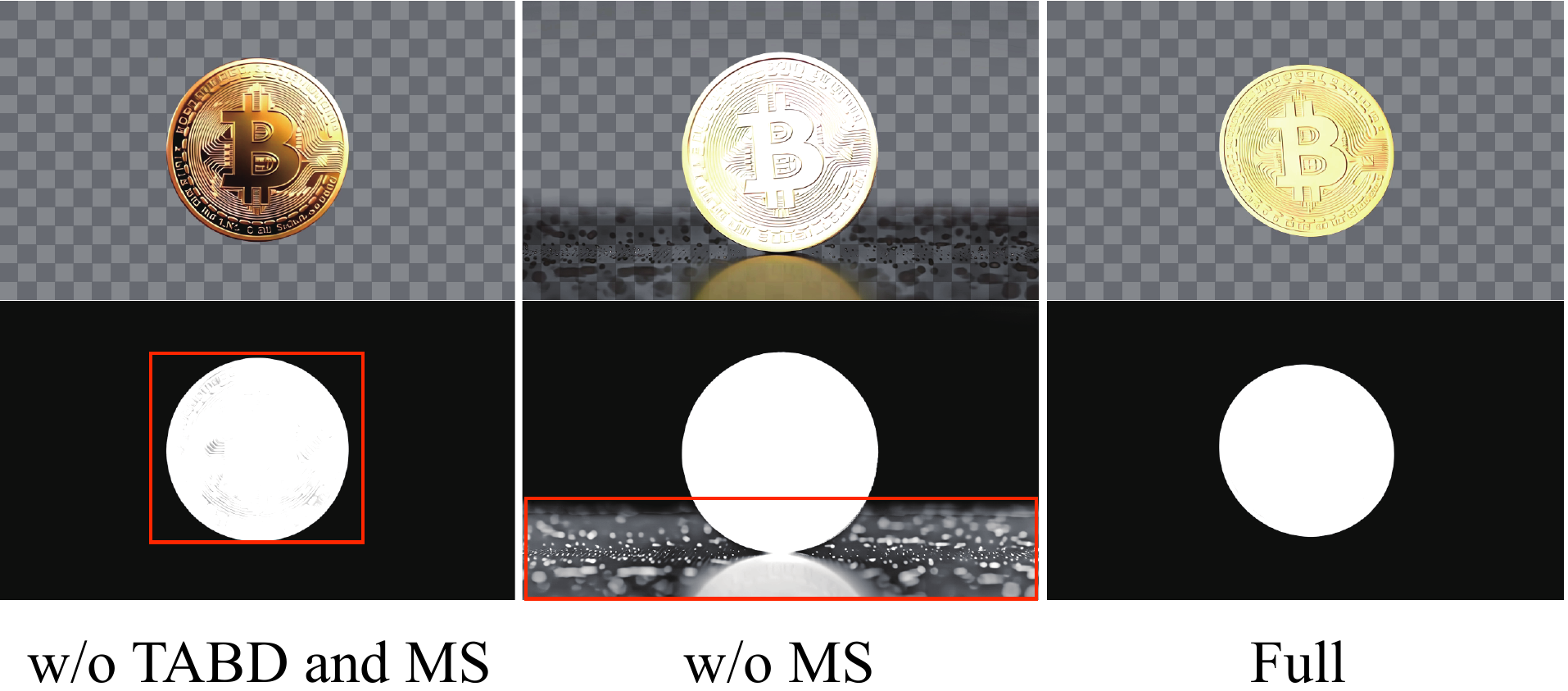}
    \vspace{-1mm}
    \caption{Ablation study on RGB-A video generation.}
    \label{fig:ablation}
    \vspace{-3mm}
\end{figure}

\noindent{\bf Analysis on VAE Design~}~We first evaluate the reconstruction performance resulting from our VAE training objective designs, as shown in Table~\ref{tab:ablation}. The evaluation uses PSNR, SSIM~\cite{wang2004image}, and LPIPS~\cite{zhang2018unreasonable}. We denote Rendering as the rendering operation employed for input pre-processing and reconstruction losses, and TABD as the transparency-aware bidirectional diffusion loss. Incorporating Rendering improves the VAE’s video reconstruction metrics, demonstrating that rendering operations help the VAE learn transparency relationships. Adding TABD further enhances reconstruction performance, as distinguishing RGB and alpha channels in the latent space facilitates their separate and more accurate reconstruction.
We illustrate the effect of TABD on video generation in Fig.~\ref{fig:ablation}. Without TABD, the video model may produce holes in opaque regions due to the VAE entangling RGB and alpha channels, making it difficult for the DiT to distinguish between them. In contrast, with TABD, the foreground content exhibits a more accurate alpha channel.

\setlength{\tabcolsep}{3.2pt}
\renewcommand{\arraystretch}{1.2}
\begin{table}[t]
    \centering
    \caption{Ablation study of RGB-A video reconstruction performance under different VAE designs. We separately present the reconstruction results for the RGB and alpha channels.}
    \vspace{-2mm}
    \label{tab:ablation}
    \footnotesize 
    \begin{tabular}{cc||cc|cc|cc}
    \hline
    \multicolumn{2}{c||}{\textbf{Method}} & \multicolumn{2}{c|}{\textbf{PSNR}$\uparrow$} & \multicolumn{2}{c|}{\textbf{SSIM}$\uparrow$} & \multicolumn{2}{c}{\textbf{LPIPS}$\downarrow$} \\\hline
    \multirow{1}{*}{\textbf{Rendering}} & \multirow{1}{*}{\textbf{TABD}}  
     & $~~$RGB$~~$ &$~~$$\alpha $$~~$ & $~~$RGB$~~$ & $~~$$\alpha$$~~$ & $~~$RGB$~~$ & $~~$$\alpha$$~~$ \\
    \hline\hline
        &  & 40.12 & 39.98 & 0.97 & 0.97 & 0.043 & 0.025 \\
    \ding{51} &  &   40.88 & 41.22 & 0.97 & 0.98 & 0.040 & 0.023 \\
    \ding{51} & \ding{51} & \textbf{41.47} & \textbf{42.22} & \textbf{0.97} & \textbf{0.98} & \textbf{0.037} & \textbf{0.022} \\
    \hline
    \end{tabular}
    \vspace{-1mm}
\end{table}

\begin{figure}[t]
    \centering
    \includegraphics[width=0.99\linewidth]{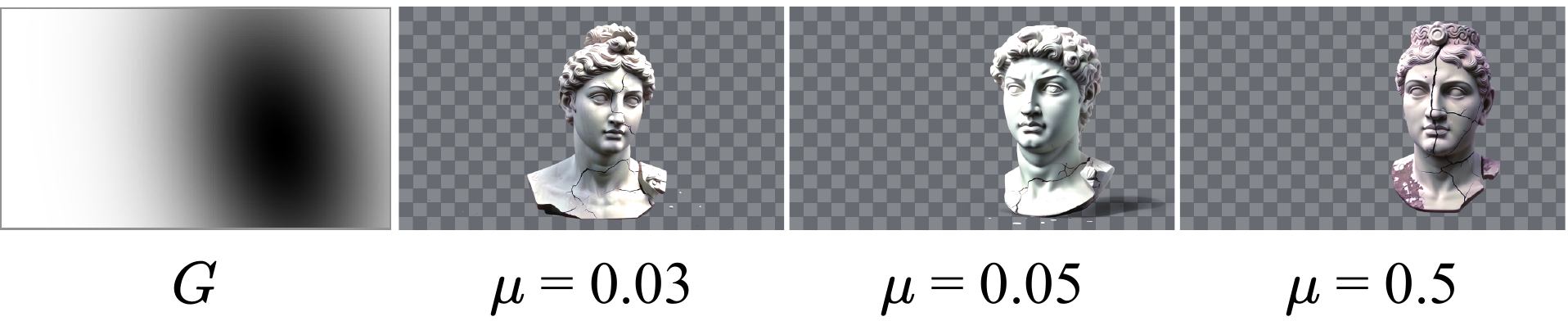}
    \vspace{-1mm}
    \caption{Effect of different $\mu$ values in our mean sampler.}
    \label{fig:mean}
\end{figure}

\vspace{1mm}
\noindent{\bf Analysis on DiT~}~Without the proposed transparency-guided mean sampler, our model cannot allow users to control the placement of transparency. Furthermore, when introducing TABD to guide the VAE, the opaque latents remain similar to those of the base model, whereas transparent latents are more challenging to learn. Consequently, the DiT may tend to generate fewer transparent regions and fails to produce clear transparent backgrounds. By contrast, our mean sampler (MS) enables a relatively robust arrangement of transparency without compromising RGB generation quality, thereby preventing unwanted background artifacts, as illustrated in Fig.~\ref{fig:ablation}. The effect of varying $\mu$ is illustrated in Fig.~\ref{fig:mean}. When $\mu$ is too small, it has little impact on transparency control. In our experiments, we typically set $\mu = 0.05$. Excessively large values (e.g., 10$\times$ higher, $\mu = 0.5$) may introduce slight red color bias.

\subsection{Application}

\begin{figure}[t]
    \centering
    \includegraphics[width=0.99\linewidth]{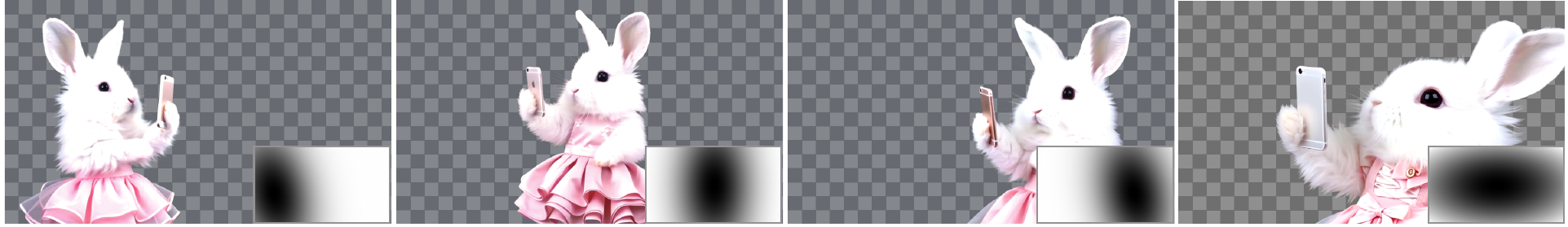}
    \vspace{-1mm}
    \caption{Results under different Gaussian ellipse masks $G$ (shown at the bottom right). Users can use $G$ to control the shape and position of transparent regions.}
    \label{fig:position}
\end{figure}

\noindent{\bf Transparency Control~}~Our transparency-guided mean sampler enables control over the approximate shape and position of transparency generation.
As shown in Fig.~\ref{fig:position}, when we move the mask, the position of the rabbit changes accordingly. When we provide a large mask, the rabbit also becomes larger. Moreover, the model automatically adjusts the rabbit’s orientation to create a more harmonious composition. This behavior arises because using a Gaussian mask provides the model with sufficient flexibility to capture fine structural details and motion. With different mask shapes, the model can automatically adapt the generated content to maintain visual naturalness, highlighting the capability and robustness of our model.

\begin{figure}[t]
    \centering
    \includegraphics[width=0.99\linewidth]{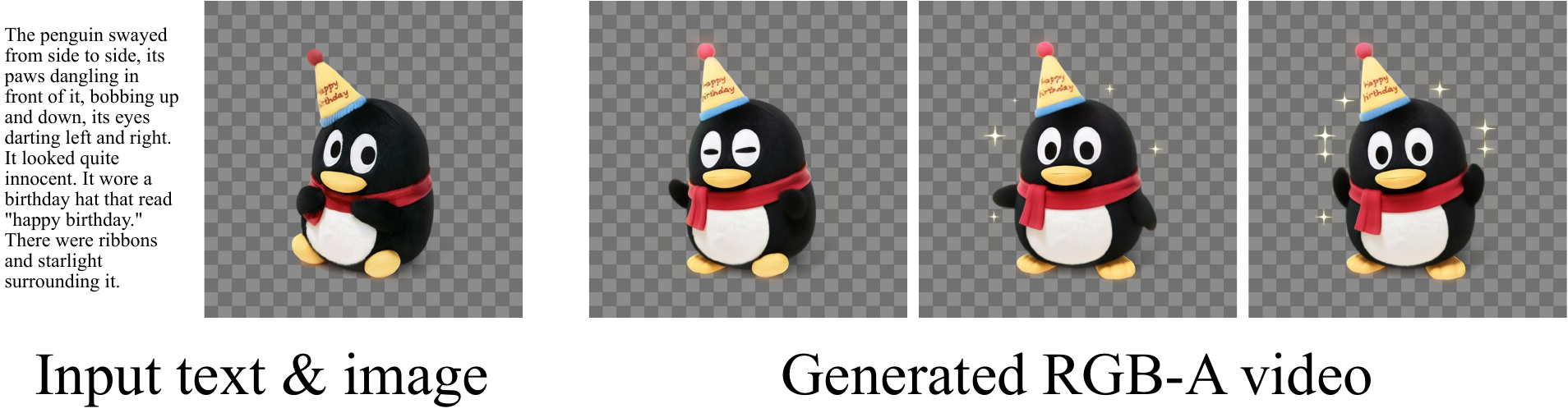}
    \vspace{-1mm}
    \caption{Results on extending to the image-to-video task, where users can animate a custom image driven by a text prompt.}
    \label{fig:IT2V}
\end{figure}

\vspace{1mm}
\noindent{\bf Image-to-Video~}~We train Wan2.1-I2V-14B~\cite{wan2025wan} using our proposed framework and dataset, enabling users to input a transparent image and animate its motion based on textual instructions as shown in Fig.~\ref{fig:IT2V}.
This functionality is especially valuable for applications such as game design.

\section{Conclusion}
\label{sec:conclusion}

We introduced a method for high‑quality RGB‑A video generation by shifting alpha distributions while preserving RGB fidelity in both latent and noise spaces. Through a transparency‑aware diffusion loss and mean‑shifted noise sampling, our approach achieves stable and controllable transparency rendering. Experiments demonstrate that our model has superior performance over state‑of‑the‑art methods in text alignment, aesthetics, naturalness, motion smoothness, and temporal consistency.
We believe this work can contribute to the advancement of the creative industry and the broader AIGC community.

{\small
\bibliographystyle{ieeenat_fullname}
\bibliography{11_references}

@String(IJCV = {Int. J. Comput. Vis.})

@String(CVPR= {IEEE Conf. Comput. Vis. Pattern Recog.})

@String(ICCV= {Int. Conf. Comput. Vis.})

@String(ECCV= {Eur. Conf. Comput. Vis.})

@String(TOG= {ACM Trans. Graph.})

@String(TIP  = {IEEE Trans. Image Process.})

@String(ICASSP=	{ICASSP})

@String(ICLR = {Int. Conf. Learn. Represent.})

@String(IJCAI = {IJCAI})

@String(AAAI = {AAAI})

@String(IJCV  = {IJCV})

@String(CVPR  = {CVPR})

@String(ICCV  = {ICCV})

@String(ECCV  = {ECCV})

@String(NIPSW  = {NeurIPS Workshop})

@String(TOG   = {ACM TOG})

@String(TIP   = {IEEE TIP})

@String(ICLR  = {ICLR})

@article{LayerDiffuse,
  author       = {Lvmin Zhang and
                  Maneesh Agrawala},
  title        = {Transparent Image Layer Diffusion using Latent Transparency},
  journal      = TOG,
  year         = {2024},
}

@article{wan2025wan,
  title={Wan: Open and advanced large-scale video generative models},
  author={Wan, Team and Wang, Ang and Ai, Baole and Wen, Bin and Mao, Chaojie and Xie, Chen-Wei and Chen, Di and Yu, Feiwu and Zhao, Haiming and Yang, Jianxiao and others},
  journal = {ArXiv preprint},
  year={2025}
}

@inproceedings{hu2022lora,
  title={Lora: Low-rank adaptation of large language models.},
  author={Hu, Edward J and Shen, Yelong and Wallis, Phillip and Allen-Zhu, Zeyuan and Li, Yuanzhi and Wang, Shean and Wang, Lu and Chen, Weizhu and others},
  booktitle={ICLR},
  year={2022}
}

@inproceedings{vgg,
  author       = {Karen Simonyan and
                  Andrew Zisserman},
  title        = {Very Deep Convolutional Networks for Large-Scale Image Recognition},
  booktitle    = {ICLR},
  year         = {2015},
}

@inproceedings{AIM,
  title     = {Deep Automatic Natural Image Matting},
  author    = {Li, Jizhizi and Zhang, Jing and Tao, Dacheng},
  booktitle = {IJCAI},
  year      = {2021},
}

@article{li2022bridging,
  title={Bridging composite and real: towards end-to-end deep image matting},
  author={Li, Jizhizi and Zhang, Jing and Maybank, Stephen J and Tao, Dacheng},
  journal={IJCV},
  year={2022},
}

@InProceedings{Qiao_2020_CVPR,
    author = {Qiao, Yu and Liu, Yuhao and Yang, Xin and Zhou, Dongsheng and Xu, Mingliang and Zhang, Qiang and Wei, Xiaopeng},
    title = {Attention-Guided Hierarchical Structure Aggregation for Image Matting},
    booktitle = {CVPR},
    year = {2020}
}

@inproceedings{sun2021dvm,
  author    = {Yanan Sun and Guanzhi Wang and Qiao Gu and Chi-Keung Tang and Yu-Wing Tai},
  title     = {Deep Video Matting via Spatio-Temporal Alignment and Aggregation},
  booktitle = {CVPR},
  year      = {2021},
}

@InProceedings{Sun_2023_CVPR,
    author    = {Sun, Yanan and Tang, Chi-Keung and Tai, Yu-Wing},
    title     = {Ultrahigh Resolution Image/Video Matting With Spatio-Temporal Sparsity},
    booktitle = {CVPR},
    year      = {2023},
}

@InProceedings{Liu_2021_ICCV,
   author    = {Liu, Yuhao and Xie, Jiake and Shi, Xiao and Qiao, Yu and Huang, Yujie and Tang, Yong and Yang, Xin},
   title     = {Tripartite Information Mining and Integration for Image Matting},
   booktitle = {ICCV},
   year      = {2021},
}

@article{rethink_p3m,
  title={Rethinking Portrait Matting with Pirvacy Preserving},
  author={Ma, Sihan and Li, Jizhizi and Zhang, Jing and Zhang, He and Tao, Dacheng},
  journal={IJCV},
  year={2023}
}

@inproceedings{yu2021mask,
  title={Mask guided matting via progressive refinement network},
  author={Yu, Qihang and Zhang, Jianming and Zhang, He and Wang, Yilin and Lin, Zhe and Xu, Ning and Bai, Yutong and Yuille, Alan},
  booktitle={CVPR},
  year={2021}
}

@inproceedings{sun2021sim,
  author    = {Yanan Sun and Chi-Keung Tang and Yu-Wing Tai},
  title     = {Semantic Image Matting},
  booktitle = {CVPR},
  year      = {2021},
}

@inproceedings{cai2022TransMatting,
  title={TransMatting: Enhancing Transparent Objects Matting with Transformers},
  author={Cai, Huanqia and Xue, Fanglei and Xu, Lele and Guo, Lili},
  booktitle={ECCV},
  year={2022},
}

@inproceedings{zhang2021attention,
  title={Attention-guided temporally coherent video object matting},
  author={Zhang, Yunke and Wang, Chi and Cui, Miaomiao and Ren, Peiran and Xie, Xuansong and Hua, Xian-Sheng and Bao, Hujun and Huang, Qixing and Xu, Weiwei},
  booktitle={ACM MM},
  year={2021}
}

@inproceedings{lin2021real,
  title={Real-time high-resolution background matting},
  author={Lin, Shanchuan and Ryabtsev, Andrey and Sengupta, Soumyadip and Curless, Brian L and Seitz, Steven M and Kemelmacher-Shlizerman, Ira},
  booktitle={CVPR},
  year={2021}
}

@inproceedings{
lipman2023flow,
title={Flow Matching for Generative Modeling},
author={Yaron Lipman and Ricky T. Q. Chen and Heli Ben-Hamu and Maximilian Nickel and Matthew Le},
booktitle={ICLR},
year={2023},
}

@inproceedings{
chung2023unimax,
title={UniMax: Fairer and More Effective Language Sampling for Large-Scale Multilingual Pretraining},
author={Hyung Won Chung and Xavier Garcia and Adam Roberts and Yi Tay and Orhan Firat and Sharan Narang and Noah Constant},
booktitle={ICLR},
year={2023},
}

@inproceedings{liu2024dora,
  title={Dora: Weight-decomposed low-rank adaptation},
  author={Liu, Shih-Yang and Wang, Chien-Yi and Yin, Hongxu and Molchanov, Pavlo and Wang, Yu-Chiang Frank and Cheng, Kwang-Ting and Chen, Min-Hung},
  booktitle={ICML},
  year={2024}
}

@inproceedings{esser2024scaling,
  title={Scaling rectified flow transformers for high-resolution image synthesis},
  author={Esser, Patrick and Kulal, Sumith and Blattmann, Andreas and Entezari, Rahim and M{\"u}ller, Jonas and Saini, Harry and Levi, Yam and Lorenz, Dominik and Sauer, Axel and Boesel, Frederic and others},
  booktitle={ICML},
  year={2024}
}

@misc{lightx2v,
 author = {LightX2V Contributors},
 title = {LightX2V: Light Video Generation Inference Framework},
 year = {2025},
 publisher = {GitHub},
 howpublished = {\url{https://github.com/ModelTC/lightx2v}},
}

@misc{qwen2.5-VL,
    title = {Qwen2.5-VL},
    howpublished = {\url{https://qwenlm.github.io/blog/qwen2.5-vl/}},
    author = {Qwen Team},
    year = {2025}
}

@article{chen2025transanimate,
      title={TransAnimate: Taming Layer Diffusion to Generate RGBA Video}, 
      author={Xuewei Chen and Zhimin Chen and Yiren Song},
      year={2025},
      journal={ArXiv preprint},
}

@inproceedings{TransVDM,
  author       = {Menghao Li and
                  Zhenghao Zhang and
                  Junchao Liao and
                  Long Qin and
                  Weizhi Wang},
  title        = {TransVDM: Motion-Constrained Video Diffusion Model for Transparent
                  Video Synthesis},
  booktitle    = {ICASSP},
  year         = {2025},
}

@inproceedings{layeranimate,
  author       = {Jingqi Bai and
                  Jingkai Zhou and
                  Benzhi Wang and
                  Weihua Chen and
                  Yang Yang and
                  Zhen Lei and
                  Fan Wang},
  title        = {Layer-Animate for Transparent Video Generation},
  booktitle    = {ICASSP},
  year         = {2025},
}

@inproceedings{AnimateDiff,
  author       = {Yuwei Guo and
                  Ceyuan Yang and
                  Anyi Rao and
                  Zhengyang Liang and
                  Yaohui Wang and
                  Yu Qiao and
                  Maneesh Agrawala and
                  Dahua Lin and
                  Bo Dai},
  title        = {AnimateDiff: Animate Your Personalized Text-to-Image Diffusion Models
                  without Specific Tuning},
  booktitle    = {ICLR},
  year         = {2024},
}

@inproceedings{ILDiff,
  author       = {Ting Zhang and
                  Zhiqiang Yuan and
                  Yeshuang Zhu and
                  Jie Zhou and
                  Jinchao Zhang},
  title        = {ILDiff: Generate Transparent Animated Stickers by Implicit Layout Distillation},
  booktitle    = {ICASSP},
  year         = {2025},
}

@inproceedings{quattrini2024alfie,
  title={Alfie: Democratising RGBA Image Generation with No},
  author={Quattrini, Fabio and Pippi, Vittorio and Cascianelli, Silvia and Cucchiara, Rita},
  booktitle={ECCV},
  year={2024},
}

@inproceedings{wang2025transpixeler,
  title={TransPixeler: Advancing Text-to-Video Generation with Transparency},
  author={Wang, Luozhou and Li, Yijun and Chen, Zhifei and Wang, Jui-Hsien and Zhang, Zhifei and Zhang, He and Lin, Zhe and Chen, Ying-Cong},
  booktitle={CVPR},
  year={2025}
}

@inproceedings{yang2025generative,
  title={Generative Image Layer Decomposition with Visual Effects},
  author={Yang, Jinrui and Liu, Qing and Li, Yijun and Kim, Soo Ye and Pakhomov, Daniil and Ren, Mengwei and Zhang, Jianming and Lin, Zhe and Xie, Cihang and Zhou, Yuyin},
  booktitle={CVPR},
  year={2025}
}

@article{huang2025psdiffusion,
  title={PSDiffusion: Harmonized Multi-Layer Image Generation via Layout and Appearance Alignment},
  author={Huang, Dingbang and Li, Wenbo and Zhao, Yifei and Pan, Xinyu and Zeng, Yanhong and Dai, Bo},
  journal={ArXiv preprint},
  year={2025}
}

@article{DreamLayer,
  author={Junjia Huang and Pengxiang Yan and Jinhang Cai and Jiyang Liu and Zhao Wang and Yitong Wang and Xinglong Wu and Guanbin Li},
  title={DreamLayer: Simultaneous Multi-Layer Generation via Diffusion Mode},
  year={2025},
journal={ArXiv preprint},
}

@article{kang2025layeringdiff,
  title={LayeringDiff: Layered Image Synthesis via Generation, then Disassembly with Generative Knowledge},
  author={Kang, Kyoungkook and Sim, Gyujin and Kim, Geonung and Kim, Donguk and Nam, Seungho and Cho, Sunghyun},
  journal={ArXiv preprint},
  year={2025}
}

@inproceedings{ji2025layerflow,
  title={LayerFlow: A Unified Model for Layer-aware Video Generation},
  author={Ji, Sihui and Luo, Hao and Chen, Xi and Tu, Yuanpeng and Wang, Yiyang and Zhao, Hengshuang},
  booktitle={ACM SIGGRAPH},
  year={2025}
}

@article{dai2025trans,
  title={Trans-Adapter: A Plug-and-Play Framework for Transparent Image Inpainting},
  author={Dai, Yuekun and Li, Haitian and Zhou, Shangchen and Loy, Chen Change},
  journal={ArXiv preprint},
  year={2025}
}

@inproceedings{perceptual_loss,
  author       = {Justin Johnson and
                  Alexandre Alahi and
                  Li Fei{-}Fei},
  title        = {Perceptual Losses for Real-Time Style Transfer and Super-Resolution},
  booktitle    = {ECCV},
  year         = {2016},
}

@article{alphavae,
  author       = {Zile Wang and
                  Hao Yu and
                  Jiabo Zhan and
                  Chun Yuan},
  title        = {AlphaVAE: Unified End-to-End {RGBA} Image Reconstruction and Generation with Alpha-Aware Representation Learning},
  journal      = {ArXiv preprint},
  year         = {2025},
}

@inproceedings{latentdiffusion,
  author       = {Robin Rombach and
                  Andreas Blattmann and
                  Dominik Lorenz and
                  Patrick Esser and
                  Bj{\"{o}}rn Ommer},
  title        = {High-Resolution Image Synthesis with Latent Diffusion Models},
  booktitle    = {CVPR},
  year         = {2022},
}

@InProceedings{huang2023vbench,
     title={{VBench}: Comprehensive Benchmark Suite for Video Generative Models},
     author={Huang, Ziqi and He, Yinan and Yu, Jiashuo and Zhang, Fan and Si, Chenyang and Jiang, Yuming and Zhang, Yuanhan and Wu, Tianxing and Jin, Qingyang and Chanpaisit, Nattapol and Wang, Yaohui and Chen, Xinyuan and Wang, Limin and Lin, Dahua and Qiao, Yu and Liu, Ziwei},
     booktitle=CVPR,
     year={2024}
 }

@article{wang2004image,
  title={Image quality assessment: from error visibility to structural similarity},
  author={Wang, Zhou and Bovik, Alan C and Sheikh, Hamid R and Simoncelli, Eero P},
  journal=TIP,
  year={2004},
}

@inproceedings{zhang2018unreasonable,
  title={The unreasonable effectiveness of deep features as a perceptual metric},
  author={Zhang, Richard and Isola, Phillip and Efros, Alexei A and Shechtman, Eli and Wang, Oliver},
  booktitle={CVPR},
  year={2018}
}

@inproceedings{Sobel1990AnI3,
  title={An Isotropic 3×3 image gradient operator},
  author={Irwin Sobel and G. M. Feldman},
  year={1990},
}

@inproceedings{CFG,
    title={Classifier-Free Diffusion Guidance},
    author={Jonathan Ho and Tim Salimans},
    booktitle=NIPSW,
    year={2021},
    url={https://openreview.net/forum?id=qw8AKxfYbI}
}

@inproceedings{MatAnyone,
  author       = {Peiqing Yang and
                  Shangchen Zhou and
                  Jixin Zhao and
                  Qingyi Tao and
                  Chen Change Loy},
  title        = {MatAnyone: Stable Video Matting with Consistent Memory Propagation},
      booktitle = {CVPR},
      year      = {2025}
}

@article{OpenS2V,
  author       = {Shenghai Yuan and
                  Xianyi He and
                  Yufan Deng and
                  Yang Ye and
                  Jinfa Huang and
                  Bin Lin and
                  Jiebo Luo and
                  Li Yuan},
  title        = {OpenS2V-Nexus: {A} Detailed Benchmark and Million-Scale Dataset for
                  Subject-to-Video Generation},
    journal = {ArXiv preprint},
  year         = {2025},
}

@article{gpt_4o,
  author       = {OpenAI},
  title        = {{GPT-4} Technical Report},
    journal = {ArXiv preprint},
  year         = {2023},
}

@inproceedings{BBDM,
  author       = {Bo Li and
                  Kaitao Xue and
                  Bin Liu and
                  Yu{-}Kun Lai},
  title        = {{BBDM:} Image-to-Image Translation with Brownian Bridge Diffusion
                  Models},
  booktitle    = {CVPR},
  year         = {2023},
}

@inproceedings{ShiftDDPMs,
  author       = {Zijian Zhang and
                  Zhou Zhao and
                  Jun Yu and
                  Qi Tian},
  title        = {ShiftDDPMs: Exploring Conditional Diffusion Models by Shifting Diffusion
                  Trajectories},
  booktitle    = AAAI,
  year         = {2023},
}

@inproceedings{Shifted_Diffusion,
  author       = {Yufan Zhou and
                  Bingchen Liu and
                  Yizhe Zhu and
                  Xiao Yang and
                  Changyou Chen and
                  Jinhui Xu},
  title        = {Shifted Diffusion for Text-to-image Generation},
  booktitle    = CVPR,
  year         = {2023},
}

@inproceedings{TKG-DM,
  author       = {Ryugo Morita and
                  Stanislav Frolov and
                  Brian Bernhard Moser and
                  Takahiro Shirakawa and
                  Ko Watanabe and
                  Andreas Dengel and
                  Jinjia Zhou},
  title        = {{TKG-DM:} Training-free Chroma Key Content Generation Diffusion Model},
  booktitle    = CVPR,
  year         = {2025},
}
}


\end{document}